\documentclass[letterpaper]{article} 
\usepackage{aaai24}  
\usepackage{times}  
\usepackage{helvet}  
\usepackage{courier}  
\usepackage[hyphens]{url}  
\usepackage{graphicx} 
\usepackage{natbib}  
\usepackage{caption} 
\usepackage{algorithm}
\usepackage{algorithmic}
\usepackage{amssymb}
\usepackage{times}
\usepackage{soul}
\usepackage{url}
\usepackage[utf8]{inputenc}
\usepackage{amsmath, amsfonts}
\usepackage{dsfont}
\usepackage{amsthm}
\usepackage{booktabs}
\usepackage{multirow}
\usepackage{subfigure}
\theoremstyle{definition}
\newtheorem{thm}{Theorem}

\newtheorem{dfn}[thm]{Definition}

\usepackage{newfloat}
\usepackage{listings}
\DeclareCaptionStyle{ruled}{labelfont=normalfont,labelsep=colon,strut=off} 
\floatstyle{ruled}
\newfloat{listing}{tb}{lst}{}
\floatname{listing}{Listing}
\title{Lyapunov-Stable Deep Equilibrium Models}
\author {
    Haoyu Chu\textsuperscript{\rm 1,\rm 2,\rm 3},
    Shikui Wei\thanks{Corresponding author.}\textsuperscript{\rm 1, \rm 3},
    Ting Liu\textsuperscript{\rm 4},
    Yao Zhao\textsuperscript{\rm 1, \rm 3},
    Yuto Miyatake\textsuperscript{\rm 5}
}
\affiliations {
    \textsuperscript{\rm 1}Institute of Information Science, Beijing Jiaotong University\\
    \textsuperscript{\rm 2}Graduate School of Information Science and Technology, Osaka University\\
    \textsuperscript{\rm 3}Beijing Key Laboratory of Advanced Information Science and Network Technology \\
    \textsuperscript{\rm 4}School of Computer Science, Northwestern Polytechnical University\\
    \textsuperscript{\rm 5}Cybermedia Center, Osaka University\\
    19112001@bjtu.edu.cn, shkwei@bjtu.edu.cn, liuting@nwpu.edu.cn, yzhao@bjtu.edu.cn, miyatake@cas.cmc.osaka-u.ac.jp
}

\usepackage{bibentry}

\begin{document}

\maketitle

\begin{abstract}
Deep equilibrium (DEQ) models have emerged as a promising class of implicit layer models, which abandon traditional depth by solving for the fixed points of a single nonlinear layer. Despite their success, the stability of the fixed points for these models remains poorly understood. By considering DEQ models as nonlinear dynamic systems, we propose a robust DEQ model named LyaDEQ with guaranteed provable stability via Lyapunov theory. The crux of our method is ensuring the Lyapunov stability of the DEQ model's fixed points, which enables the proposed model to resist minor initial perturbations. To avoid poor adversarial defense due to Lyapunov-stable fixed points being located near each other, we orthogonalize the layers after the Lyapunov stability module to separate different fixed points. We evaluate LyaDEQ models under well-known adversarial attacks, and experimental results demonstrate significant improvement in robustness. Furthermore, we show that the LyaDEQ model can be combined with other defense methods, such as adversarial training, to achieve even better adversarial robustness.
\end{abstract}

\section{Introduction}

Deep equilibrium models have demonstrated remarkable progress in various deep learning tasks, such as language modeling, image classification, semantic segmentation, compressive imaging, and optical flow estimation~\cite{bai2020multiscale,winston2020monotone, zhao2023deep, bai2022deep}. Unlike conventional neural networks that rely on stacking layers, DEQ models define their outputs as solutions to an input-dependent fixed points equation and use arbitrary black-box solvers to reach the fixed points without storing intermediate activations. As a result, DEQ models are categorized as implicit networks, presenting a unique approach to deep learning.

However, the robustness of DEQ models remains largely unexplored. As widely known, deep neural networks (DNNs) are susceptible to adversarial examples, which are crafted with minor perturbations to input images. Given the pervasive use of deep learning in various aspects of daily life, the emergence of adversarial examples poses a severe threat to the security of deep learning systems~\cite{szegedy2013intriguing, lu2023black, zhang2023improving}. Hence, it is imperative to investigate the robustness of DEQ models. An intriguing question arises: can adversarial examples easily deceive DEQ models as well? If so, can we fundamentally mitigate this issue?

\citet{wei2021certified} showed that DEQ models are also vulnerable to adversarial examples and considered $\ell_{\infty}$ certified robustness for DEQ models. They presented IBP-MonDEQ, a modification of monotone deep equilibrium layers that allows for the computation of lower and upper bounds on its output via interval bound propagation. Nevertheless, our experimental analysis revealed that IBP-MonDEQ does not provide significant improvement in adversarial robustness for some complex image recognition tasks. Likewise, \citet{li2022cerdeq} proposed a defense method for DEQ models based on certified training. 

The deep learning community has shown a great interest in improving the adversarial robustness of neural networks. For another kind of implicit network, neural ordinary differential equations (Neural ODEs)~\cite{chen2018neural}, defense methods based on the Lyapunov method have emerged owing to their connection with dynamical systems. \citet{kang2021stable} proposed a stable Neural ODE with Lyapunov-stable equilibrium points for defending against adversarial attacks. \citet{rodriguez2022lyanet} proposed a method for training ODEs by using a control-theoretic Lyapunov condition for stability, leading to improvement on adversarial robustness. 

Typically, a stable dynamical system implies that all solutions in some region around an equilibrium point (i.e., in a neighborhood of an equilibrium point) flow to that point. Lyapunov's direct method generalizes this concept by reasoning about convergence to states that minimize a potential Lyapunov function. Because Lyapunov theory deals with the effect of the initial perturbations on dynamic systems, integrating Lyapunov theory into implicit layers can automatically confer many benefits, such as adversarial robustness.

 \begin{figure*}[t]
\begin{center}
\includegraphics[width=0.98\textwidth]{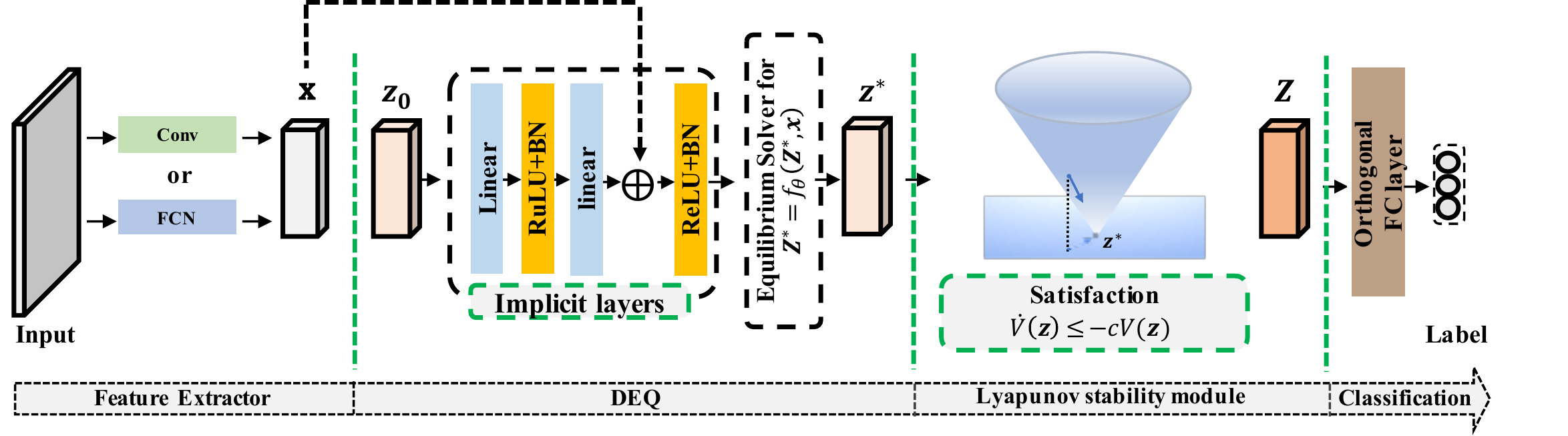}
\end{center}
\caption{
The scratch of the architecture of the LyaDEQ model. The blue arrow represents a state that locally satisfies the Lyapunov exponential stability condition.}
\label{fig:overview}
\end{figure*}

In this paper, we present a novel approach for improving the robustness of DEQ models through provable stability guaranteed by the Lyapunov theory. Unlike existing methods that rely on certified training or adversarial training \cite{wei2021certified, li2022cerdeq, yang2022closer, yang2023improving}, our approach treats the DEQ model as a nonlinear dynamic system and ensures that its fixed points are Lyapunov-stable, thereby keeping the perturbed fixed point within the same stable neighborhood as the unperturbed point and preventing successful adversarial attacks. Specially, we ensure the robustness of the DEQ model by jointly learning a convex positive definite Lyapunov function along with dynamics constrained to be stable according to these dynamics everywhere in the state space. Consequently, the minor adversarial perturbations added to the input image will hardly change the output of the DEQ model. Besides, for classification problems, Lyapunov-stable fixed points for different classes may be located near each other, leading to each stable neighborhood being very small, resulting in poor robustness against adversarial examples. To address this issue, we propose to use orthogonalization techniques to increase the distance between Lyapunov stable equilibrium points. The architecture of the LyaDEQ model is shown in Figure~\ref{fig:overview}. 

We name our proposed model LyaDEQ. Our main contributions are summarized as follows:

(1) We introduce Lyapunov stability theory into the DEQ model by considering it as a nonlinear system, which enables us to certify the stability of the fixed points. To the best of our knowledge, this is the first attempt to utilize the Lyapunov stability framework in the DEQ model.

(2) To address the poor adversarial defense caused by the small stable neighborhood of the fixed points, we introduce an orthogonal parametrized fully connected (FC) layer after the Lyapunov stability module to separate different Lyapunov-stable fixed points.

(3) The experimental results on MNIST, Street View House Numbers (SVHN), and CIFAR10/100 datasets demonstrate that the proposed LyaDEQ model consistently outperforms the baseline model in terms of robustness against adversarial attacks. These results validate the applicability of the Lyapunov theory to match the DEQ model and support the correctness of our theoretical analysis.

(4) We show that the LyaDEQ model can be combined with other adversarial training methods such as TRADES~\cite{zhang2019theoretically}, robust dataset~\cite{ilyas2019adversarial}, and PGD-AT~\cite{madry2017towards}, to achieve even better adversarial robustness. 


\section{Related Works}

This section reviews works related to deep equilibrium models and Lyapunov theory in deep learning.

\subsection{Deep Equilibrium Models}

Motivated by an observation that the hidden layers of many existing deep sequence models converge towards some fixed points, DEQ models~\cite{bai2019deep} find these fixed points via the root-finding method. Due to DEQ models suffering from unstable convergence to a solution and lacking guarantees that a solution exists, \citet{winston2020monotone} proposed a monotone operator equilibrium network, which guarantees stable convergence to a unique fixed point. \citet{bai2020multiscale} proposed the multi-scale deep equilibrium model for handling large-scale vision tasks, such as ImageNet classification and semantic segmentation on high-resolution images. Later, they presented a regularization scheme for DEQ models that explicitly regularizes the Jacobian of the fixed-point update equations to stabilize the learning of equilibrium models~\cite{bai2021stabilizing}. Furthermore, \citet{bai2021neural} introduced neural deep equilibrium solvers for DEQ models to improve the speed/accuracy trade-off across diverse large-scale tasks. \citet{li2021optimization} proposed the multi-branch optimization-induced equilibrium models based on modeling the hidden objective function for the multi-resolution recognition task. \citet{tsuchida2021declarative} showed that solving a kernelized regularised maximum likelihood estimate as an inner problem in a deep declarative network yields a large class of DEQ architectures. \citet{tsuchida2023deep} presented a DEQ model that solves the problem of joint maximum a-posteriori estimation in a graphical model representing nonlinearly parameterized exponential family principal component analysis. \citet{gilton2021deep} presented an approach based on DEQ models for solving the linear inverse problems in imaging.

\subsection{Lyapunov Theory in Deep Learning}

Lyapunov functions are convenient tools for the stability certification of dynamical systems. Recently, many researchers have leveraged the Lyapunov stability theory to construct provable, neural network-based safety certificates. \citet{kolter2019learning} used a learnable (i.e., defined by neural network architectures) Lyapunov function to modify a base dynamics model to ensure the stability of equilibrium. \citet{richards2018lyapunov} constructed a neural network Lyapunov function and a training algorithm to adapt them to the shape of the largest safe region for a closed-loop dynamical system. \citet{chang2019antisymmetricrnn} proposed to use anti-symmetric weight matrices to parametrize an RNN from the Lyapunov stability perspective, which enhances its long-term dependency.

Since the appearance of Neural ODEs, integrating Lyapunov methods into Neural ODEs has become a new trend. Inspired by LaSalle’s theorem (an extension of Lyapunov's direct method), \citet{takeishi2021learning} proposed a deep dynamics model that can handle the stability of general types of invariant sets such as limit cycles and line attractors. They used augmented Neural ODEs~\cite{dupont2019augmented} as the invertible feature transform for the provable existence of a stable invariant set. \citet{massaroli2020stable} introduced stable neural flows whose trajectories evolve on monotonically non-increasing level sets of an energy functional parametrized by a neural network. Based on classical time-delay stability theory, \citet{schlaginhaufen2021learning} proposed a new regularization term based on a neural network Lyapunov–Razumikhin function to stabilize neural delay differential equations.

\section{Preliminaries}

For a nonlinear system $\frac{\mathrm d}{\mathrm dt}\boldsymbol{u} = F(\boldsymbol{u})$, a state $\boldsymbol{u^{\star}}$ is a fixed point (or an equilibrium point) of a nonlinear system if $\boldsymbol{u^{\star}}$ satisfies $F(\boldsymbol{u^{\star}}) = \boldsymbol{0}$. A nonlinear system can have several (or infinitely many) isolated fixed points. One of the common interests in analyzing dynamical systems is the Lyapunov stability of the fixed points. A fixed point is stable means that the trajectories starting near $\boldsymbol{u^{\star}}$ remain around it all the time. More formally;


\begin{dfn}[Lyapunov stability]
    \label{def1}
    An equilibrium $\boldsymbol{u^{\star}}$ is said to be stable in the sense of Lyapunov, if for every $\varepsilon>0$, there exists $\delta>0$ such that, if $\|\boldsymbol{u}(0)-\boldsymbol{u^{\star}}\|<\delta$, then $\|\boldsymbol{u}(t)-\boldsymbol{u^{\star}}\|<\varepsilon$ for all $t \geq 0$. If $\boldsymbol{u^{\star}}$ is stable, and $\lim_{t \rightarrow \infty}\Vert \boldsymbol{u}(t) - \boldsymbol{u^{\star}}\Vert = 0 $, $\boldsymbol{u^{\star}}$ is said to be asymptotically stable. If $\boldsymbol{u^{\star}}$ is stable and for $\nu>0$, if $\lim_{t \rightarrow \infty}\Vert \boldsymbol{u}(t) - \boldsymbol{u^{\star}}\Vert e^{\nu t}= 0$, $\boldsymbol{u^{\star}}$ is said to be exponentially stable.
\end{dfn}

\begin{thm}[Lyapunov stability theorem]~\cite{giesl2015review}
    \label{thm}
    Let $\boldsymbol{u^{\star}}$ be a fixed point. Let $V: \mathcal{U} \rightarrow \mathds{R}$ be a continuously differentiable function, defined on a neighborhood $\mathcal{U}$ of $\boldsymbol{u^{\star}}$, which satisfies
\begin{enumerate}
    \item[(1)] 
    $V$ has a minimum at $\boldsymbol{u^{\star}}$. A sufficient condition is $V(\boldsymbol{u}) \geq 0$ for all $\boldsymbol{u} \in \mathcal{U}$ and $V(\boldsymbol{u})=$ $0 \Leftrightarrow \boldsymbol{u}=\boldsymbol{u^{\star}}$.
    \item[(2)] 
    $V$ is strictly decreasing along solution trajectories of $F$ in $\mathcal{U}$ except for the fixed point. A sufficient condition is $\dot{V}(\boldsymbol{u})<0$ for all $\boldsymbol{u} \in$ $\mathcal{U} \backslash\{\boldsymbol{u^{\star}}\}$, where 
    \begin{equation}
    \dot{V}({\boldsymbol{u}}) = \frac{dV}{dt} = \nabla V(\boldsymbol{u})^TF(\boldsymbol{u})<0.
    \end{equation}
\end{enumerate}
If such a function $V$ exists, then it is called a Lyapunov function, and $\boldsymbol{u^{\star}}$ is \textit{asymptotically stable}. Moreover, $\boldsymbol{u^{\star}}$ is \textit{exponentially stable} if there exists positive definite $V$ and some $K(\alpha) > 0$ such that 

    \begin{enumerate}
    \item[(3)]  
    $ \| \boldsymbol{u} \|^2_2 \leq V(\boldsymbol{u}) \leq K(\alpha) \| \boldsymbol{u} \|^2_2 $, for all $\boldsymbol{u}$ that $\| \boldsymbol{u} \| \leq \alpha $; 
    \item[(4)] 
    $\dot{V}({\boldsymbol{u}})  \leq -cV(\boldsymbol{u})$, $c>0$.
    \end{enumerate}
\end{thm}

\section{Methodology}

In this section, we first provide a dynamic system perspective for the DEQ model. Then, we introduce the Lyapunov stability framework, which is essential to our proposed model. Finally, we present the LyaDEQ model as a novel framework for enhancing the robustness of the DEQ model.

\subsection{Considering a DEQ Model as a Nonlinear System}

Given an input $\boldsymbol{x}$, a DEQ model~\cite{bai2019deep} aims to specify a layer $f_{\theta}$ that finds the fixed points of the following iterative procedure
\begin{equation}
\boldsymbol{z}_{i+1}=f_\theta(\boldsymbol{z}_{i}, \boldsymbol{x}),
\end{equation}
where $i=0, \ldots, L-1$. Usually, we set $\boldsymbol{z}_{0} = \boldsymbol{0}$ and choose the layer $f_{\theta}$ as a shallow block, such as a fully connected layer and convolutional layer. 

Unlike a conventional neural network where the outputs are the activations from the $L^{th}$ layer, the outputs of a DEQ model are the fixed points. One can alternatively find the fixed points $\boldsymbol{z}^{\star}=f_\theta(\boldsymbol{z}^{\star}, \boldsymbol{x})$ directly via root-finding algorithms rather than fixed points iteration alone:
\begin{equation}
f_\theta(\boldsymbol{z}^{\star}, \boldsymbol{x})-\boldsymbol{z}^{\star}=0.
\end{equation}

Efficient root-finding algorithms, such as Broyden’s method~\cite{broyden1965class} and Anderson acceleration~\cite{anderson1965iterative}, can be applied to find this solution.

By defining $F({\boldsymbol{z^{\star}}}) = f_\theta(\boldsymbol{z}^{\star}, \boldsymbol{x})-\boldsymbol{z}^{\star}=0$, we consider a DEQ model as a nonlinear dynamical system, i.e., a nonlinear system parameterized by a DEQ model. By doing so, we can use Lyapunov's theory to stabilize the fixed points and enable the DEQ model to resist minor initial perturbations on the inputs. 

\subsection{Lyapunov Stability Framework}

Lyapunov's direct method is a powerful tool for studying the stability of dynamical systems. It aims to determine whether a system's final state, influenced by initial perturbations, can return to its original equilibrium state. Asymptotic stability, as defined in Definition \ref{def1}, means that any initial state near the equilibrium state will eventually approach the equilibrium state. Exponential stability, on the other hand, ensures that the system's trajectory decays at a minimum attenuation rate.

In this paper, we consider the fixed points of the DEQ model as the equilibrium state and mainly focus on the adversarial perturbations added to the input images. We aim to ensure the stability of the deep equilibrium models by jointly learning a convex Lyapunov function along with dynamics constrained to be stable according to these dynamics everywhere in the state space. As a result, our proposed LyaDEQ model is anticipated to exhibit robustness against adversarial examples.

We use neural networks to learn a positive definite Lyapunov function $V$ that satisfies conditions (1) and (3) in Theorem \ref{thm} and project outputs of a base dynamics model onto a space where condition (4) also holds~\cite{kolter2019learning}. 

Let $F(\boldsymbol{z}^{\star}): \mathds{R}^n \rightarrow \mathds{R}^n$ be a basic dynamic system parametrized by a DEQ model, let $V: \mathds{R}^n \rightarrow \mathds{R}$ be a positive definite function, and $\alpha$ be a nonnegative constant, the Lyapunov-stable nonlinear dynamic model is defined as

\begin{equation}
\begin{aligned}
\hat{F}(\boldsymbol{z}^{\star}) & =\operatorname{Proj}(F(\boldsymbol{z}^{\star}),\{F: \nabla V(\boldsymbol{z}^{\star})^T F \leq-\alpha V(\boldsymbol{z}^{\star})\}) \\
&= \begin{cases}F(\boldsymbol{z}^{\star}) & \text { if } \phi(\boldsymbol{z}^{\star}) \leq 0, \\ 
F(\boldsymbol{z}^{\star})-\nabla V(\boldsymbol{z}^{\star}) \frac{\phi(\boldsymbol{z}^{\star})}{\|\nabla V(\boldsymbol{z}^{\star})\|_2^2} & \text {otherwise },\end{cases} 
\end{aligned}
\label{eqn: model}
\end{equation}
where $\phi(\boldsymbol{z}^{\star})= \nabla V(\boldsymbol{z}^{\star})^T F(\boldsymbol{z}^{\star}) +\alpha V(\boldsymbol{z}^{\star})$.

The Lyapunov function $V$ is defined as positive definite and continuously differentiable, and has no local minima:
\begin{equation}
V(\boldsymbol{z}^{\star})=\sigma_{k+1}(g(\boldsymbol{z}^{\star})-g(0))+\|\boldsymbol{z}^{\star}\|_2^2,
\end{equation}
where $\sigma_{k}$ is a positive convex non-decreasing function with $\sigma_{k}(0) = 0$, and $g$ is represent as an input-convex neural network (ICNN)~\cite{amos2017input}:
\begin{equation}
\begin{aligned}
\boldsymbol{q}_1 &=\sigma_0\left(W^{I}_0 \boldsymbol{z}^{\star}+b_0\right), \\
\boldsymbol{q}_{i+1} &=\sigma_i\left(U_i \boldsymbol{q}_i+W^{I}_i \boldsymbol{z}^{\star}+b_i\right), i=1, \ldots, k-1, \\
g(\boldsymbol{z}^{\star}) & \equiv \boldsymbol{q}_k,
\end{aligned}
\end{equation}
where $W^{I}_i$ are real-valued weights, $b_i$ are real-valued biases, and $U_i$ are positive weights.

\noindent \textbf{Proposition 1} The function $V$ is convex in $\boldsymbol{z}^{\star}$ provided that all $U_i$ are non-negative, and all functions $\sigma_{i}$ are convex and non-decreasing.

\begin{proof}
The proof follows from the fact that non-negative sums of convex functions are convex and that the composition of a convex and convex non-decreasing function is also convex. For more detailed proof, please refer to~\cite{boyd2004convex}.
\end{proof}

Through the procedures described above, we can therefore ensure that the DEQ model satisfies the conditions of the Lyapunov stability theorem. As a result, the fixed points of the modified DEQ model become exponentially stable. 

\subsection{LyaDEQ Model}

As shown in Figure~\ref{fig:overview}, our proposed model, LyaDEQ, consists of a feature extractor, a DEQ model, a Lyapunov stability module, and an orthogonal FC layer. 

\subsubsection{Feature extractor}

The feature extractor plays the role of dimensionality reduction. In our experiment, we choose a fully connected network (FCN) or ResNet~\cite{he2016deep} as the backbone of LyaDEQ. 

\subsubsection{DEQ}  

A DEQ model ultimately finds the fixed points of a single function $\boldsymbol{z}^{\star} = f_\theta(\boldsymbol{z}^{\star}, \boldsymbol{x})$. We define the implicit layer $f_\theta$ as a feed-forward neural network, which can be written formally as 

\begin{equation}
\begin{aligned}
\boldsymbol{y} =  W^{D}_{2} \text{BN} (\text{ReLU} (W^{D}_1 \boldsymbol{z}_0 + b) + b \\
f_\theta(\boldsymbol{z}, \boldsymbol{x}) = \text{BN} (\text{ReLU} ( \boldsymbol{x} + \boldsymbol{y} )),
\end{aligned}
\label{eqn::implicit}
\end{equation}
where $W^{D}$ are real-valued weights and $\text{BN}$ represents the batch normalization operator. 

We use Anderson acceleration~\cite{anderson1965iterative} to find the fixed points of the DEQ model.

\subsubsection{Lyapunov stability module} 

\begin{figure}[t]
\begin{center}
\includegraphics[width=0.3\textwidth]{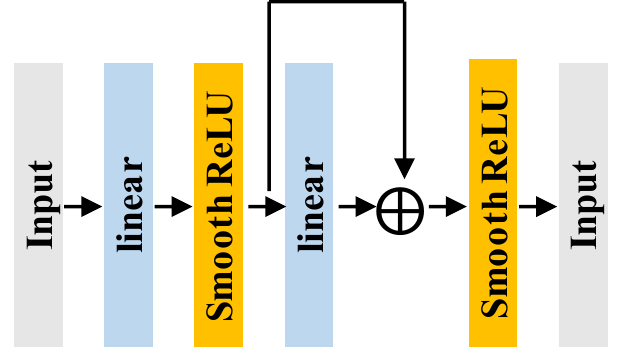}
\end{center}
\caption{The architecture of our used ICNN.}
\label{fig:architecuture}
\end{figure}

We define ICNN as a 2-layer fully connected neural network. The network architecture is shown in Figure~\ref{fig:architecuture}.  The activation function $\sigma$ is chosen as a smooth ReLU function: 

\begin{equation}
\sigma(x)= \begin{cases}0 & \text { if } x \leq 0, \\ x^2 / 2 d & \text { if } 0<x<d, \\ x-d / 2 & \text { otherwise.}\end{cases}
\end{equation}

\subsubsection{Orthogonal FC layer}

As one can see from Definition \ref{def1}, Lyapunov stability is established within a smaller stable neighborhood. For classification problems, Lyapunov-stable fixed points for different classes may be very close to each other, leading to each stable neighborhood may be very small, resulting in poor robustness against adversarial examples (see the experimental results on MNIST in Table \ref{tab::main}). 

\begin{figure*}[t]
\centering
\resizebox{\linewidth}{!}{
\subfigure[LyaDEQ (clean)]{
\begin{minipage}[t]{0.25\linewidth}
\centering
\includegraphics[width=1in]{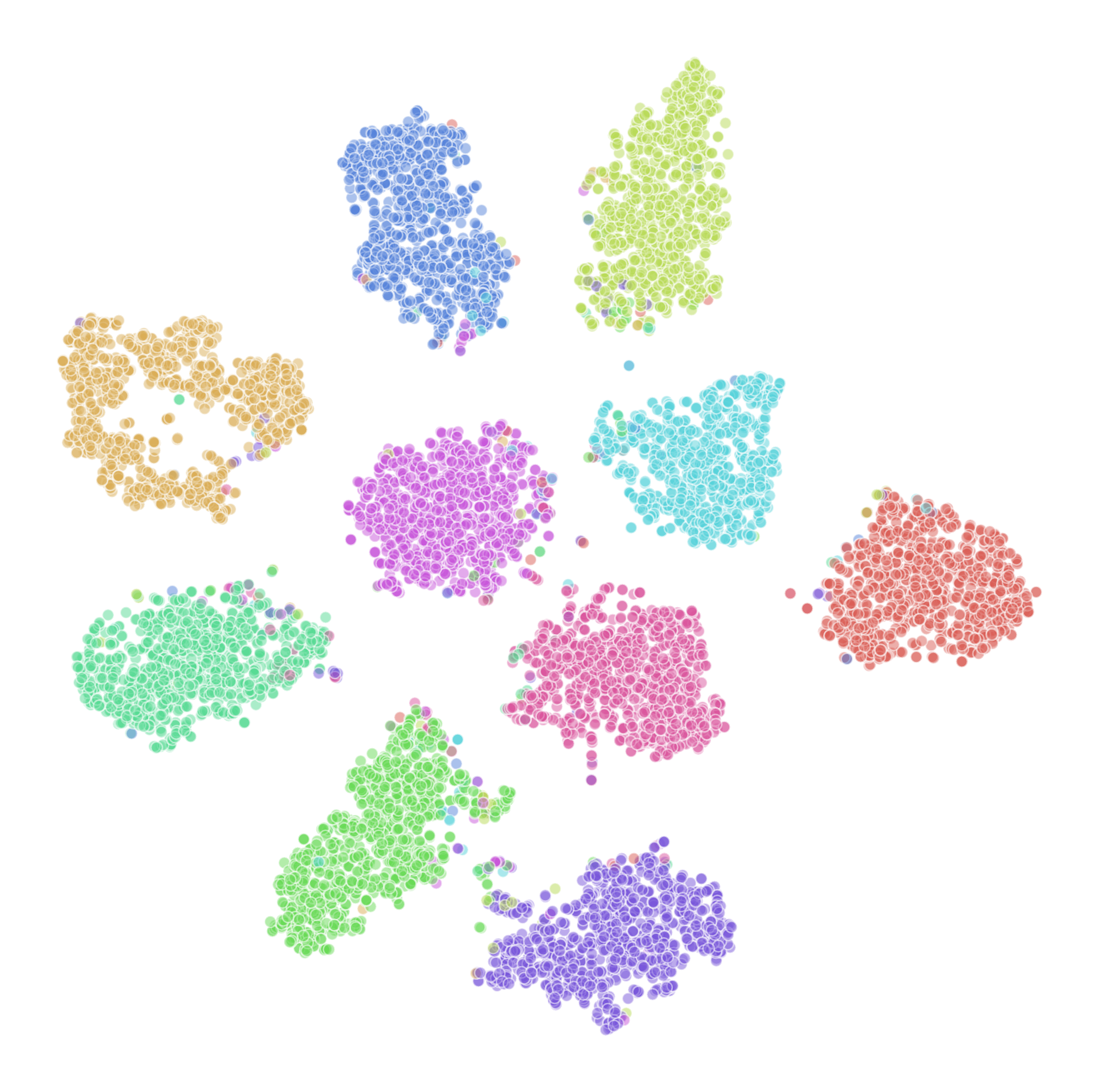}
\end{minipage}%
}%
\subfigure[LyaDEQ w/o orthog. FC (clean)]{
\begin{minipage}[t]{0.25\linewidth}
\centering
\includegraphics[width=1in]{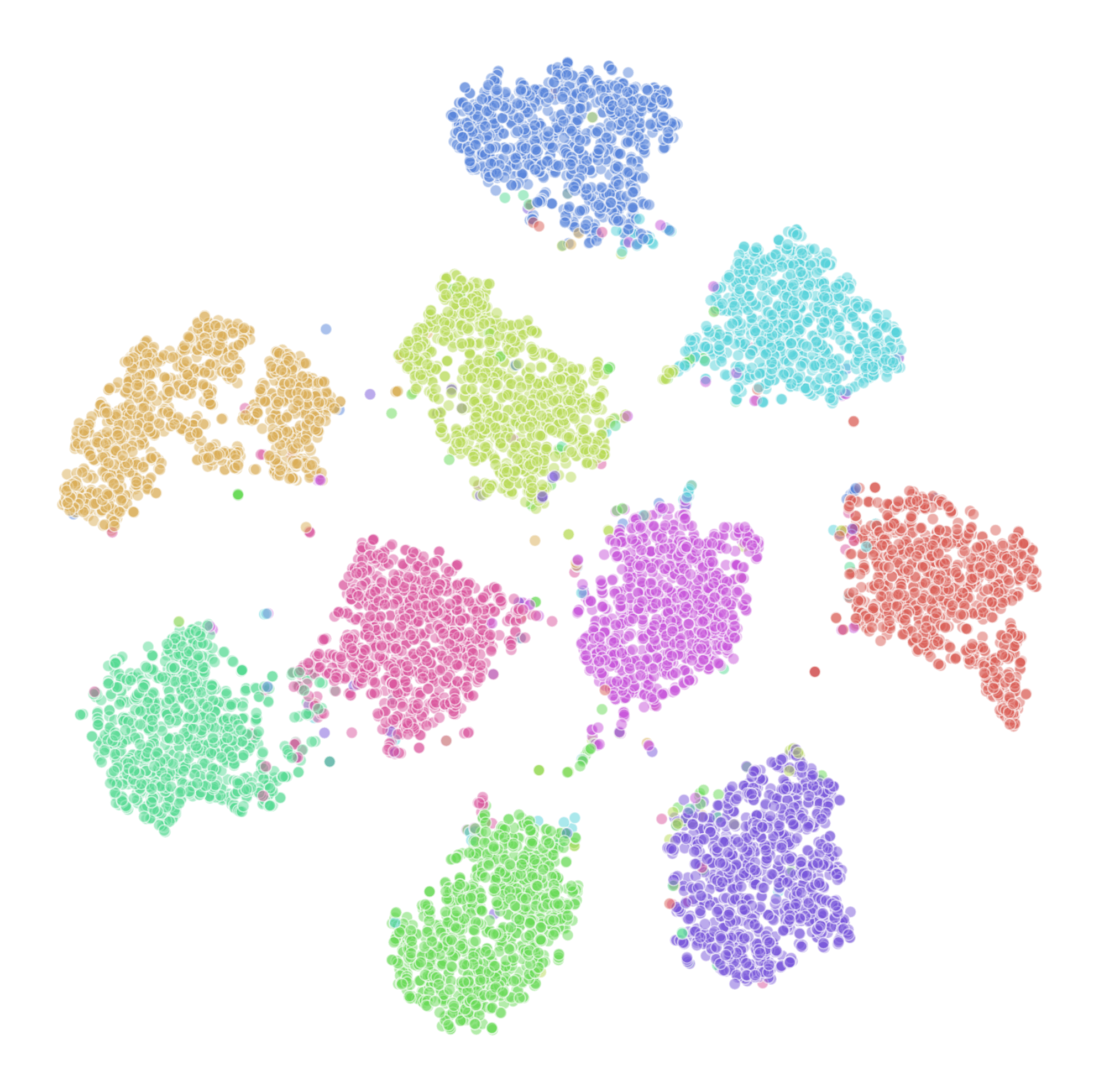}
\end{minipage}%
}%
\subfigure[LyaDEQ (adv)]{
\begin{minipage}[t]{0.25\linewidth}
\centering
\includegraphics[width=1in]{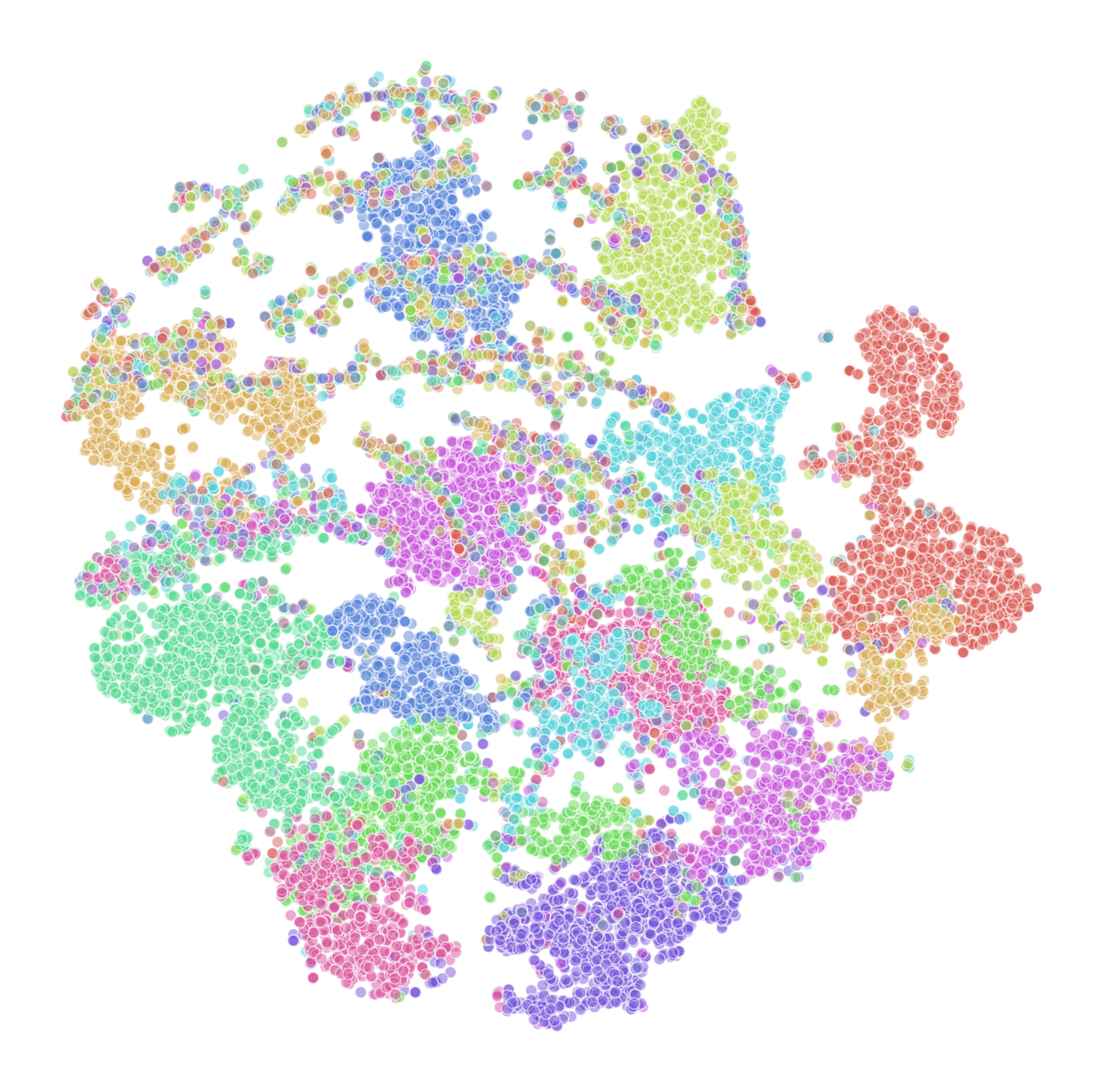}
\end{minipage}
}%
\subfigure[LyaDEQ w/o orthog. FC (adv)]{
\begin{minipage}[t]{0.25\linewidth}
\centering
\includegraphics[width=1in]{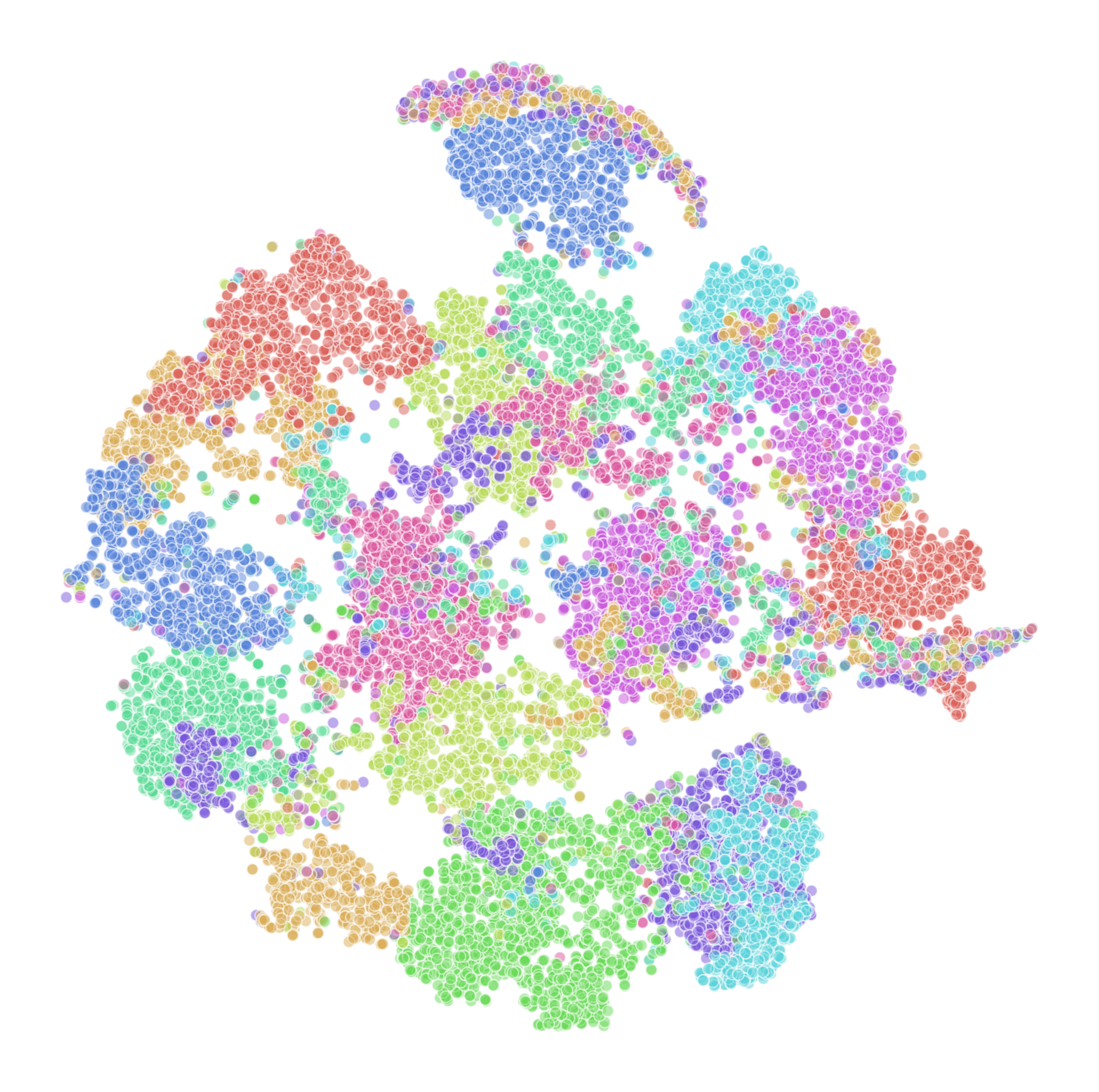}
\end{minipage}
}%
}
\centering
\caption{t-SNE visualization results on the features after the orthogonal FC layer. `adv' means test on the adversarial dataset.}
\label{fig:tsne}
\end{figure*}

We add an orthogonal FC layer after the Lyapunov stability module to increase the distance between Lyapunov stable equilibrium points. Given the output of the Lyapunov stability module $\boldsymbol{Z}$, the orthogonal FC layer will return the parametrized version $\boldsymbol{Z}$ so that $\boldsymbol{Z}^{T}\boldsymbol{Z}=\boldsymbol{I}$. The t-SNE visualization of the features after the orthogonal FC layer is shown in Figure \ref{fig:tsne}. 

\begin{table*}[t] 
\centering
\resizebox{\linewidth}{!}{
\begin{tabular}{cccccccccc}
\hline
Benchmark                 & Model                                         & Clean                  & Attack  & $\epsilon=2/255$        & $\epsilon=4/255$        & $\epsilon=6/255$        & $\epsilon=8/255$        \\ \hline
\multirow{10}{*}{MNIST}    & \multirow{2}{*}{DEQ (baseline)}               & \multirow{2}{*}{96.99} & I-FGSM & 50.24                   & 49.89                   & 49.67                   & 49.34                   \\
                          &                                               &                        & PGD    & 45.98                   & 45.80                   & 45.47                   & 45.23                   \\ \cline{2-8} 
                          & \multirow{2}{*}{DEQ w/ orthog. FC(ablation)}    & \multirow{2}{*}{96.56} & I-FGSM & 29.98                   & 29.91                   & 29.60                   & 29.44                   \\
                          &                                               &                        & PGD    & 29.94                   & 29.94                   & 29.71                   & 29.53                   \\	\cline{2-8} 
                          & \multirow{2}{*}{IBP-MonDEQ}    & \multirow{2}{*}{99.29} & I-FGSM &      \textbf{96.52}              &         \textbf{96.38}          &        \textbf{96.32}            &     \textbf{96.27}              \\
                          &                                               &                        & PGD    &    \textbf{94.87}              &     \textbf{94.80}             &      \textbf{94.69}           &    \textbf{94.57}         \\	\cline{2-8} 
                          & \multirow{2}{*}{LyaDEQ w/o orthog. FC (ours)} & \multirow{2}{*}{97.10} & I-FGSM & 23.18                   & 23.28                   & 23.21                   & 23.25                   \\
                          &                                               &                        & PGD    & 23.17                   & 23.26                   & 23.24                   & 23.23                   \\\cline{2-8} 
                          & \multirow{2}{*}{LyaDEQ w/ orthog. FC (ours)}  & \multirow{2}{*}{96.59} & I-FGSM & \underline{50.78} (+0.54)  & \underline{50.60} (+0.71)  & \underline{50.55} (+0.88)  & \underline{50.43} (+1.09)  \\
                          &                                               &                        & PGD    & \underline{50.72} (+4.74)  & \underline{50.54} (+4.74)  & \underline{50.52} (+5.05)  & \underline{50.35} (+5.12) \\  \hline

\multirow{10}{*}{SVHN}     & \multirow{2}{*}{DEQ (baseline)}               & \multirow{2}{*}{95.37} & I-FGSM & 68.09                   & 61.78                   & 56.77                   & 51.65                   \\
                          &                                               &                        & PGD    & 67.06                   & 60.89                   & 56.38                   & 51.22                   \\ \cline{2-8} 
                          & \multirow{2}{*}{DEQ w/ orthog. FC (ablation)} & \multirow{2}{*}{95.63} & I-FGSM & 67.30                   & 59.96                   & 54.27                   & 48.79                   \\
                          &                                               &                        & PGD    & 66.74                   & 60.31                   & 55.59                   & 50.02                   \\ \cline{2-8} 
                          & \multirow{2}{*}{IBP-MonDEQ}    & \multirow{2}{*}{91.06} & I-FGSM &   57.71                 &   55.71                &    53.78                &    51.58               \\
                          &                                               &                        & PGD    &   57.75               &    55.74              &   54.15             &      52.04       \\	\cline{2-8} 
                          & \multirow{2}{*}{LyaDEQ w/o orthog. FC (ours)} & \multirow{2}{*}{95.28} & I-FGSM & \textbf{72.16 (+4.07)}  & \textbf{72.08 (+10.30)} & \textbf{71.84 (+15.07)} & \textbf{71.42 (+19.77)} \\
                          &                                               &                        & PGD    & \underline{68.95}                   & \underline{67.88}                   & \underline{67.11}                   & \underline{65.60}                   \\ \cline{2-8} 
                          & \multirow{2}{*}{LyaDEQ w/ orthog. FC (ours)}  & \multirow{2}{*}{95.21} & I-FGSM & \underline{69.41}                   & \underline{69.01}                   & \underline{68.55}                   & \underline{67.81}                   \\
                          &                                               &                        & PGD    & \textbf{71.35 (+4.29)}  & \textbf{71.23 (+10.34)} & \textbf{70.95 (+14.57)} & \textbf{70.47 (+19.25)} \\ \hline

\multirow{10}{*}{CIFAR10}  & \multirow{2}{*}{DEQ (baseline)}               & \multirow{2}{*}{87.71} & I-FGSM & 34.25                   & 23.00                   & 16.65                   & 12.36                   \\
                          &                                               &                        & PGD    & 33.37                   & 22.87                   & 16.81                   & 11.12                   \\ \cline{2-8} 
                          & \multirow{2}{*}{DEQ w/ orthog. FC (ablation)} & \multirow{2}{*}{87.62} & I-FGSM & 35.46                   & 23.48                   & 16.91                   & 11.61                   \\
                          &                                               &                        & PGD    & 32.67                   & 22.07                   & 15.92                   & 10.89                   \\ \cline{2-8} 
                          & \multirow{2}{*}{IBP-MonDEQ}    & \multirow{2}{*}{80.25} & I-FGSM &  25.87          &    24.54               &    23.59             &       22.24            \\
                          &                                               &                        & PGD    &    27.59              &     26.21             &   25.04              &      23.58       \\	\cline{2-8} 
                          & \multirow{2}{*}{LyaDEQ w/o orthog. FC (ours)} & \multirow{2}{*}{87.80} & I-FGSM & \underline{43.80}                   & \underline{43.99}                   & \underline{43.75}                   & \underline{43.45}                   \\
                          &                                               &                        & PGD    & \textbf{47.09 (+13.72)} & \textbf{46.94 (+24.07)} & \textbf{46.70 (+29.89)} & \textbf{46.12 (+35.00)} \\ \cline{2-8} 
                          & \multirow{2}{*}{LyaDEQ w/ orthog. FC (ours)}  & \multirow{2}{*}{87.87} & I-FGSM & \textbf{47.38 (+13.13)} & \textbf{47.31 (+24.31)} & \textbf{46.71 (+30.06)} & \textbf{45.80 (+33.44)} \\
                          &                                               &                        & PGD    & \underline{45.04}                   & \underline{44.81}                   & \underline{44.50}                   & \underline{43.96}                   \\ \hline

\multirow{10}{*}{CIFAR100} & \multirow{2}{*}{DEQ (baseline)}               & \multirow{2}{*}{61.23} & I-FGSM & 15.62                   & 8.59                    & 5.94                    & 4.15                    \\
                          &                                               &                        & PGD    & 11.85                   & 6.71                    & 4.47                    & 3.10                    \\ \cline{2-8} 
                          & \multirow{2}{*}{DEQ w/ orthog. FC (ablation)} & \multirow{2}{*}{61.41} & I-FGSM & 14.75                   & 8.13                    & 5.59                    & 3.85                    \\
                          &                                               &                        & PGD    & 13.81                   & 7.70                    & 5.16                    & 3.39                    \\ \cline{2-8} 
                           & \multirow{2}{*}{IBP-MonDEQ}    & \multirow{2}{*}{44.78} & I-FGSM &     9.15               &    8.32               &       7.68             &      7.04             \\
                          &                                               &                        & PGD    &     8.88             &      8.21            &    7.56             &      7.04       \\	\cline{2-8} 
                          & \multirow{2}{*}{LyaDEQ w/o orthog. FC (ours)} & \multirow{2}{*}{60.52} & I-FGSM & \underline{22.20}                   & \underline{22.16}                   & \underline{21.77}                   & \underline{21.07}                   \\
                          &                                               &                        & PGD    & \underline{20.25}                   & \underline{19.78}                   & \underline{19.10}                   & \underline{18.18}                   \\ \cline{2-8} 
                          & \multirow{2}{*}{LyaDEQ w/ orthog. FC (ours)}  & \multirow{2}{*}{61.30} & I-FGSM & \textbf{23.82 (+8.20)}  & \textbf{23.76 (+15.17)} & \textbf{23.73 (+17.79)} & \textbf{23.35 (+19.20)} \\
                          &                                               &                        & PGD    & \textbf{21.86 (+10.01)} & \textbf{21.53 (+14.82)} & \textbf{20.65 (+16.18)} & \textbf{19.55 (+16.45)}  \\ \hline
\end{tabular}}
\caption{Classification accuracy on MNIST, SVHN, and CIFAR. Results that surpass all competing methods are bold. The second best result is with the underline. The performance gain in parentheses is compared with the baseline model. The input is the test set of CIFAR10. We abbreviate the LyaDEQ model without the orthogonal FC layer as LyaDEQ w/o orthog. FC.}
\label{tab::main}
\end{table*}

\section{Experiments}

In this section, we first present the experimental setup. We then proceed to evaluate the performance of the LyaDEQ model against two white-box adversarial attacks. Subsequently, we demonstrate the effectiveness of the LyaDEQ model trained with adversarial training, comparing it to conventional convolutional neural networks (CNNs) in terms of robustness. Lastly, we conduct an ablation study to investigate the role of the orthogonal FC layer.

\subsection{Setup}

We conduct a set of experiments on three standard datasets MNIST~\cite{lecun1998gradient}, CIFAR10/100~\cite{krizhevsky2009learning}, and SVHN~\cite{netzer2011reading}

\subsubsection{Training configurations}

We use PyTorch~\cite{paszke2017automatic} framework for the implementation. For MNIST, we use 1-layer FCN as the feature extractor to reduce the dimension from 28$\times$28 to 64. For SVHN and CIFAR10, we use ResNet20~\cite{he2016deep} to reduce the dimension from 32$\times$32$\times$3 to 64 (128 for CIFAR100). 


For optimization, we use Adam algorithm~\cite{kingma2014adam} with betas=(0.9, 0.999). We set the initial learning rate to 0.001 and set the learning rate of each parameter group using a cosine annealing schedule. The training epochs for MNIST, SVHN, and CIFAR are set to 10, 40, and 50.

\subsubsection{The configurations of adversarial attacks} 

We test the performance of the original DEQ model and our proposed LyaDEQ model on two white-box adversarial attacks: iterative fast gradient sign method (I-FGSM) ~\citep{kurakin2018adversarial} and project gradient descent (PGD) \citep{madry2017towards}.

\noindent \textbf{I-FGSM}\ As an iterative version based FGSM~\cite{goodfellow2014explaining}, I-FGSM computes an adversarial example by multiple gradients:
\begin{equation}
\boldsymbol{x}_{n+1}^{adv}=\operatorname{Clip}_{x, \epsilon}\{\boldsymbol{X}_n^{adv}+\alpha \operatorname{sign}(\nabla_x \mathcal{L}(\boldsymbol{x}_n^{adv}, y))\},
\end{equation}
where $\alpha$ is the step size, $\operatorname{Clip}_{x, \epsilon}$ means clipping perturbed images within $[x-\epsilon,x+\epsilon]$, and $\boldsymbol{x}_{0}^{adv}=\boldsymbol{x}$. 

\noindent \textbf{PGD} \ attack is a universal attack utilizing the local first order information about the network:
\begin{equation}
\boldsymbol{x}^{adv}_{i+1}=\Pi_{\boldsymbol{x}+\mathcal{S}}\left(\boldsymbol{x}^{adv}_{i}+\alpha \operatorname{sign}\left(\nabla_{\boldsymbol{x}} \mathcal{L}_{\boldsymbol{w}}(\boldsymbol{x}, \boldsymbol{y})\right)\right),
\end{equation}
where $\Pi_{\boldsymbol{x}+\mathcal{S}}$ represents the projection on $\epsilon$-ball, $\mathcal{S}\subseteq \mathbb{R}^d$. A uniform random noise is first added to the clean image $\boldsymbol{x}$, that is $\boldsymbol{x}^{adv}_{0} = \boldsymbol{x}+ \mathcal{U} [-\epsilon, \epsilon]$. We set the size of perturbation $\epsilon$ of PGD in the infinite norm sense.

For both PGD and I-FGSM, the step size $\alpha $ is set to 1/255, and the number of steps $n$ is calculated as $n = \lfloor\text{min}(\epsilon \cdot 255 + 4, \epsilon \cdot 255 \cdot 1.25)\rfloor$.

\subsection{The Robustness of LyaDEQ Model Against Adversarial Examples}

Table \ref{tab::main} displays the results of our experiments regarding classification accuracy and robustness against adversarial examples. Regarding classification accuracy on clean data, our proposed model achieves comparable performance with the baseline model.

Regarding robustness against adversarial examples, we evaluate the effectiveness of our proposed LyaDEQ model in defending against white-box attacks with attack radii ranging from $\epsilon= 2/255$ to $\epsilon= 8/255$. Our experimental results demonstrate that the LyaDEQ model outperforms the baseline model on each dataset. For instance, compared with the DEQ model under PGD attack with $\epsilon= 8/255$, the LyaDEQ model exhibits a 5.12\%, 19.25\%, 32.48\%, and 16.45\% improvement on MNIST, SVHN, CIFAR10, and CIFAR100 datasets, respectively. These findings confirm that the Lyapunov stability module can significantly enhance the robustness of the DEQ model.

The results presented in Table \ref{tab::main} demonstrate that the magnitude of the accuracy boost under adversarial attack increases with increasing attack radii for all datasets. For instance, LyaDEQ model under I-FGSM with attack radii $\epsilon= 2/255$, $\epsilon= 4/255$, $\epsilon= 6/255$ and $\epsilon= 8/255$ has a 13.13\%, 24.31\%, 30.06\%,  33.44\% boost respectively on CIFAR10. This further corroborates the effectiveness of our proposed approach.

In addition, while the experimental results of the IBP-MonDEQ model on MNIST are much better than our method, it performs poorly on SVHN, CIFAR10, and CIFAR100 datasets. We consider that the reason for conflict lies in the complexity of image datasets. For MNIST, the scene of the image is quite simple. In contrast, for SVHN and CIFAR, the scene of the image is more complex. Thus, our proposed LyaDEQ model is better suited for handling complex image recognition tasks and can be widely utilized in commonly used datasets.

\noindent \textbf{Explanations on performance improvement:} The core concept of our proposed method revolves around utilizing the Lyapunov direct method to ensure the stability of the fixed points of the DEQ models. According to the Lyapunov stability theory, if the magnitude of perturbations on $\boldsymbol{z}$ exceeds the stable neighborhood, the impact on the final outcome becomes unpredictable, resulting in potential misclassification by the model. Conversely, if the magnitude of perturbations remains within the stable neighborhood, the final outcome remains unaffected, enabling our model to effectively resist adversarial noise and maintain robustness against adversarial attacks.

\begin{table}[t] 
\centering
\begin{tabular}{ccccc}
\hline
Radius                            & Attack                  & +TRADES & +RD    &+PAT\\ \hline
\multirow{2}{*}{$\epsilon= 2/255$} & I-FGSM   &    \textbf{72.81}           &      51.96      & \underline{60.46}            \\
                                  & PGD      & 48.48         & \underline{52.00}      &   \textbf{60.43}          \\ \hline
\multirow{2}{*}{$\epsilon= 4/255$} & I-FGSM   &     \textbf{72.68}          &    51.78   & \underline{60.34}                \\
                                  & PGD    & 48.39          & \underline{51.85 }        &   \textbf{60.42}        \\ \hline
\multirow{2}{*}{$\epsilon= 6/255$} & I-FGSM  &   \textbf{72.66}            &    51.62   & \underline{60.18}               \\
                                  & PGD    & 48.35         & \underline{51.54 }       &    \textbf{60.41}         \\ \hline
\multirow{2}{*}{$\epsilon= 8/255$} & I-FGSM   &   \textbf{72.42}            &     51.51  & \underline{60.08}                \\
                                  & PGD      & 48.33  & \underline{51.63 }      &     \textbf{60.23}        \\ \hline
\end{tabular}
\centering
\caption{Classification accuracy of the LyaDEQ model combined with adversarial training method on CIFAR10 under adversarial attacks.}
\label{tab::advtrain}
\end{table}

\begin{table}[] 
\small
\centering
\begin{tabular}{cccccccc}
\hline
Radius                            & Attack & ResNet56  & VGG16   & WResNet      \\ \hline
\multirow{2}{*}{$\epsilon= 2/255$} & I-FGSM &  56.79      & 58.63  & 54.41    \\
                                  & PGD    &  52.70   & 55.90  &  51.58   \\ \hline
\multirow{2}{*}{$\epsilon= 4/255$} & I-FGSM &  49.35       & 51.50  & 46.05     \\
                                  & PGD    & 45.07  &   49.24  & 43.74       \\ \hline
\multirow{2}{*}{$\epsilon= 6/255$} & I-FGSM & 44.88              & 45.60  & 41.43     \\
                                  & PGD    & 41.35    & 43.70   & 39.68       \\ \hline
\multirow{2}{*}{$\epsilon= 8/255$} & I-FGSM & 40.51           & 37.94  & 37.08    \\
                                  & PGD    & 36.58   & 36.46  & 35.11     \\ \hline                               
\end{tabular}
\centering
\caption{Classification accuracy of the conventional CNNs on CIFAR10 under adversarial attacks.}
\label{tab::conventional}
\end{table}

\subsection{LyaDEQ Model With Adversarial Training}

Our method is orthogonal to other adversarial defense methods, such as adversarial training, which means we can combine the LyaDEQ model with adversarial training to achieve further defense performance. We choose three commonly used adversarial training methods, TRADES~\cite{zhang2019theoretically}, robust dataset (RD)~\cite{ilyas2019adversarial} and PGD-AT (PAT)~\cite{madry2017towards}. 

\noindent \textbf{TRADES} is a defense method to trade adversarial robustness off against accuracy via combining tricks of warmup, early stopping, weight decay, batch size, and other hyperparameter settings. In our experiments, we set  perturbation epsilon = 0.031, perturbation step size = 0.007, number of iterations = 10, beta = 6.0 on the training dataset.

\noindent \textbf{RD} is created by removing non-robust features from the dataset, which yields good robust accuracy on the unmodified test set.

\noindent \textbf{PAT} is a defense method to inject adversarial examples that generated by the PGD attack into training data. It is worth mentioning that \citet{yang2022closer} also used PAT to train DEQ models. In our experiments, we set  perturbation epsilon = 0.031, perturbation step size = 0.00784, and number of iterations = 7 on the training dataset.

From Table \ref{tab::advtrain}, we see that training the LyaDEQ model with adversarial training methods can further improve robustness against adversarial examples. For example, training the LyaDEQ model with TRADES, the robust dataset and PAT show a 25.43\%, 4.58\%, and 13.08\% boost respectively on CIFAR10 under I-FGSM attack with $\epsilon= 2/255$. 

\begin{figure*}[t] 
\centering
\begin{center}
\subfigure[Under I-FGSM attacks]{\includegraphics[scale=0.28]{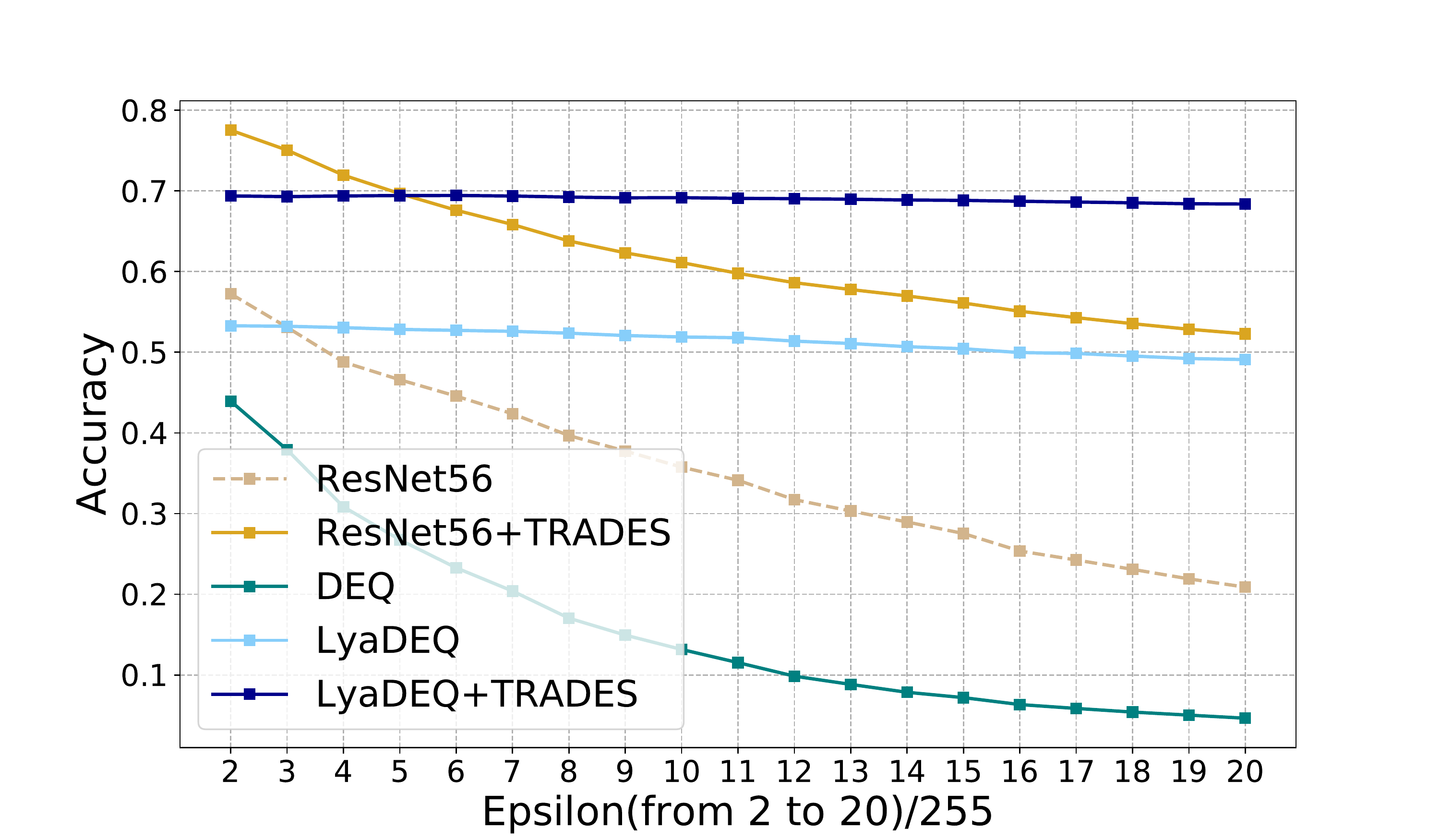}
  \label{1}}
\quad
\subfigure[Under PGD attacks]{\includegraphics[scale=0.28]{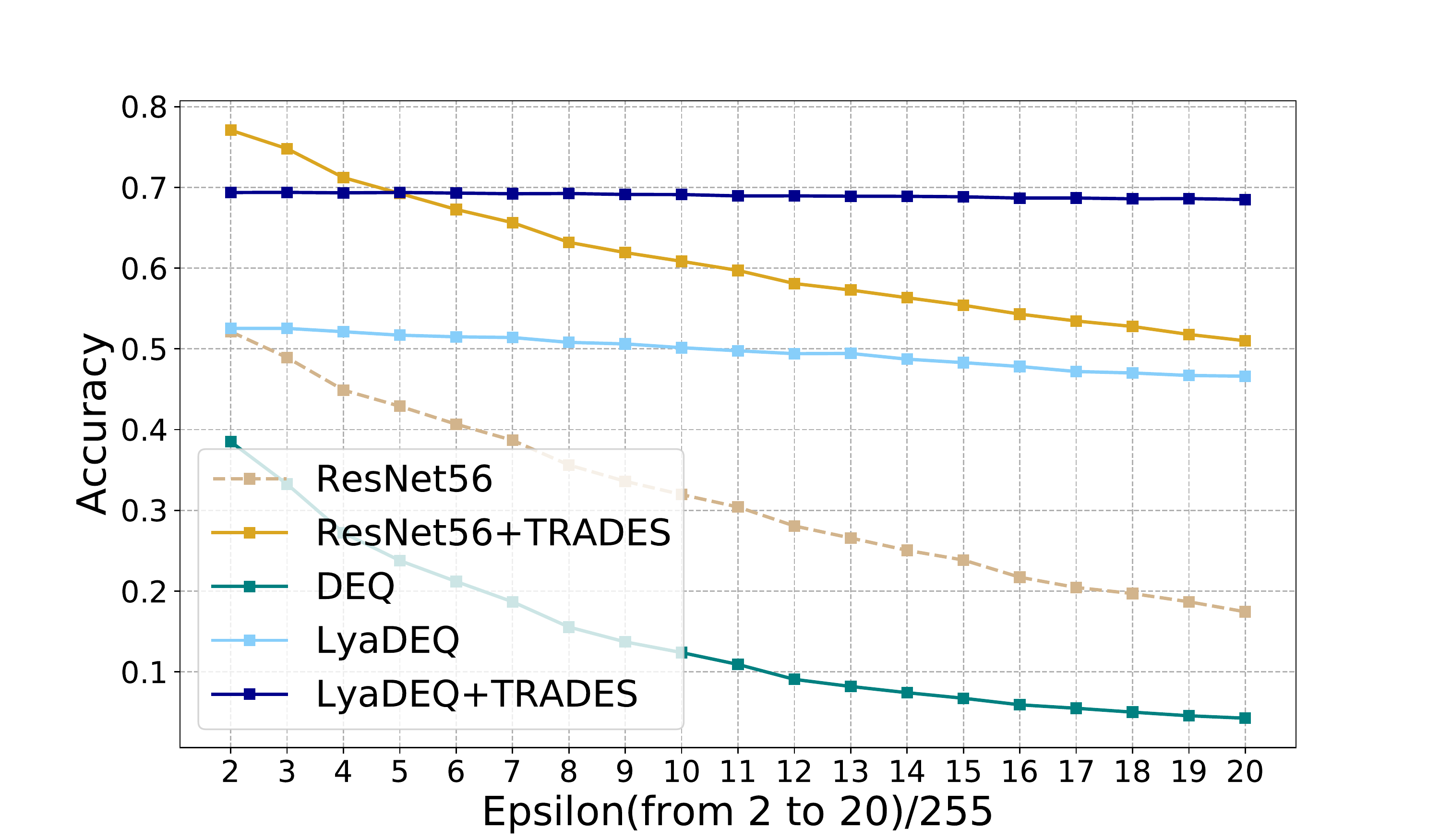}
 \label{2}}
\caption{Comparison between the DEQ model, the LyaDEQ model, and conventional CNNs on CIFAR10.}
\label{fig:comparison}
\end{center}
\end{figure*}

\subsection{Comparison With the Conventional CNNs}

We perform experiments to compare the robustness of our proposed LyaDEQ model with three conventional CNNs, ResNet56~\cite{he2016deep}, VGG16~\cite{simonyan2014very}, and WideResNet (WResNet)~\cite{zagoruyko2016wide} (with depth = 28, widen factor = 10). Table \ref{tab::conventional} summarizes the experimental results. While the robustness of the LyaDEQ model is inferior to that of conventional CNNs when the value of $\epsilon$ is small (e.g., when $\epsilon=2/255, 4/255$), the robustness of the LyaDEQ model is significantly better when the value of $\epsilon$ is getting larger (e.g., when $\epsilon=6/255, 8/255$) than these neural networks. For instance, under the PGD attack with $\epsilon=8/255$, the accuracy of the LyaDEQ model is 7.38\%, 7.50\%, and 8.85\% higher than that of ResNet56, VGG16, and WResNet, respectively.

This is because the conventional CNNs are very sensitive to the radius of the adversarial attacks. Increasing the radius has a significant impact on these networks, when the magnitude is increased from $\epsilon=2/255$ to $\epsilon=8/255$, the accuracy of ResNet56, VGG16, and WideResNet under I-FGSM attacks decreases by 16.28\%, 20.69\%, and 17.33\%, respectively. Whereas, the accuracy of the LyaDEQ model only decreases by 1.58\% in the same case, which shows that the LyaDEQ model is insensitive to the radius of the adversarial attack. Its insensitivity to the radius of adversarial attack becomes more evident when the value of $\epsilon$ is much larger. Figure \ref{fig:comparison} provides a detailed comparison between DEQ, LyaDEQ, LyaDEQ trained by TRADES, ResNet56, and ResNet trained by TRADES under adversarial attacks ranging from $\epsilon=2/255$ to $\epsilon=20/255$, further validating the robustness of the LyaDEQ model. 

\subsection{An Ablation Study}

As an ablation study, we test the LyaDEQ model without the orthogonal FC layer and the DEQ model with the orthogonal FC Layer. As we expected, adding the orthogonal FC layer to the DEQ model does not have a significant impact. From Table \ref{tab::main}, we find that in most cases, the orthogonal FC layer indeed plays a part in improving the accuracy of the LyaDEQ model under adversarial attack. Especially, on the MNIST dataset, the absence of the orthogonal FC layer incurred a substantial 20\% loss in accuracy, underscoring its crucial contribution to robustness. 

In addition, we notice the orthogonal FC layer occasionally incurs a minor deleterious effect. Nevertheless, this negative impact is acceptable when juxtaposed against the overall robustness improvement brought by the LyaDEQ model. 

\section{Conclusions}

Inspired by Lyapunov stability theory, we introduced a provably stable variant of DEQ models. Our proposed model consists of a feature extractor, a DEQ model, a Lyapunov stability module, and an orthogonal FC layer. The Lyapunov stability module ensures the fixed points of the DEQ model are Lyapunov stable, and the orthogonal FC layer separates different Lyapunov-stable fixed points. Our findings highlighted the proposed method in improving the robustness of the DEQ model.


\section{Acknowledgments}
This work was supported in part by the National Key R\&D Program of China (No.2021ZD0112100), the National Natural Science Foundation of China (No.62106201, No.61972022, No.U1936212, No.62120106009), the China Scholarship Council (No.202207090082), JSPS KAKENHI (20H01822, 20H00581, 21K18301), and JST PRESTP (JPMJPR2129).

\bibliography{aaai24}

\end{document}